\documentclass[10pt,twocolumn,letterpaper]{article}

\usepackage{cvpr}              

\usepackage{graphicx}
\usepackage{amsmath}
\usepackage{amssymb}
\usepackage{booktabs}

\newcommand*{\affaddr}[1]{#1} 
\newcommand*{\affmark}[1][*]{\textsuperscript{#1}}

\usepackage[pagebackref,breaklinks,colorlinks]{hyperref}

\usepackage[capitalize]{cleveref}
\crefname{section}{Sec.}{Secs.}
\Crefname{section}{Section}{Sections}
\Crefname{table}{Table}{Tables}
\crefname{table}{Tab.}{Tabs.}

\def\cvprPaperID{*****} 
\def\confName{CVPR}
\def\confYear{2022}

\begin{document}

\title{Low-Resolution Action Recognition for Tiny Actions Challenge in ActivityNet}

\author{Boyu Chen\affmark[1]\affmark[2], 
Yu Qiao\affmark[1]\affmark[3]\footnotemark[1],
Yali Wang\affmark[1]\affmark[4]\footnotemark[1]\\ 
\affaddr{\affmark[1] ShenZhen Key Lab of Computer Vision and Pattern Recognition, \\
Shenzhen Institute of Advanced Technology, 
Chinese Academy of Sciences, China}\\
\affaddr{\affmark[2] University of Chinese Academy of Sciences}\\
\affaddr{\affmark[3] Shanghai AI Laboratory, Shanghai, China}\\
\affaddr{\affmark[4] SIAT Branch, Shenzhen Institute of Artificial Intelligence and Robotics for Society}\\
{\tt\small \{chenboyu18@mails.ucas.ac.cn, yl.wang@siat.ac.cn, yu.qiao@siat.ac.cn\}}\\
}

\maketitle

\renewcommand{\thefootnote}{\fnsymbol{footnote}} 
\footnotetext[1]{Yali Wang (yl.wang@siat.ac.cn) and Yu Qiao (yu.qiao@siat.ac.cn) are equally-contributed corresponding authors.}

\begin{abstract}
Tiny Actions Challenge focuses on understanding human activities in real-world surveillance.
Basically,
there are two main difficulties for activity recognition in this scenario.
First,
human activities are often recorded at a distance,
and appear in a small resolution without much discriminative clue.
Second,
these activities are naturally distributed in a long-tailed way.
It is hard to alleviate data bias for such heavy category imbalance.
To tackle these problems,
we propose a comprehensive recognition solution in this paper.
First,
we train video backbones with data balance,
in order to alleviate overfitting in the challenge benchmark.
Second,
we design a dual-resolution distillation framework,
which can effectively guide low-resolution action recognition by super-resolution knowledge.
Finally,
we apply model ensemble with post-processing,
which can further boost performance on the long-tailed categories.
\textbf{Our solution ranks Top-1 on the leaderboard.}

\end{abstract}

\section{Introduction}

\label{intro}

Video action recognition is an important problem in computer vision~\cite{chen2025g, chen2025lvagent, chen2025percept, chen2025super, chen2025top, chen2025videochat, chen2025vragent, yue2025uniflow, wang2025videochat}.
However,
current approaches mainly work on high-quality videos,
which are often limited to recognize low-resolution human activities in practice.
To investigate this problem,
researchers often manually create low-resolution videos,
by down-sampling  high-resolution ones.
However,
such low-resolution videos cannot reflect real-world video quality.
To fill this gap,
Tiny Actions Challenge introduces a low-resolution activity recognition task,
where
the benchmark is collected from real surveillance videos.
Basically,
this task contains two main difficulties. 
First,
human activities are often recorded at a distance,
which is far from surveillance camera.
Hence,
human often appear in a small resolution of video,
with blurred appearance and motion information.
Second,
these activities are collected from the real life.
Hence,
their data distribution is long-tailed with heavy class imbalance.


To tackle these difficulties in this challenge,
we propose a comprehensive solution for low-resolution action recognition in videos.
Specifically,
there are three key contributions in our main solution.
First,
we choose suitable video backbones for this task,
and train them with data balance to reduce model overfitting.  \cite{li2022uniformer}
Second,
we design a dual-resolution distillation framework to boost low-resolution action recognition,
with complementary knowledge from super-resolution network.
Finally,
we perform model ensemble with post-processing,
which can further improve recognition performance on long-tail categories.


\section{Video Backbones with Data Balance}
\label{training}

In this challenge,
we mainly choose two fundamental video backbones,
including 
ir-CSN-ResNet152 \cite{tran2019video} and UniFormer-Base\cite{li2022uniformer},
which are pretrained on Kinetics400.
The main reason is that,
these models share similar advantages of 
simple structure, 
light computation, 
less overfitting, 
and 
better accuracy on spatial-temporal representation learning.
However,
directly fine-tuning these models on the competition benchmark may not be preferable,
since 
these low-resolution videos have limited action details and follow the long-tailed distribution \cite{demir2021tinyvirat}.

To have a good start,
we choose distinct settings for training models with these tiny action videos.
First,
in both the training and testing phase, 
we uniformly divided one video into 16 clips,
and randomly selected one frame in each clip. 
We find that,
this sampling setting works pretty well for F1-score improvement,
with sufficient long-range action contexts in the video.
Second,
due to long-tailed distribution in this challenge,
we propose to operate data balance in the training phase.
Specifically,
we first apply horizontal flip on the training videos of long-tail categories.
Then,
we use these flipped videos as extra training samples in these categories.  
As shown in Table \ref{tab:training},
all these settings can boost video backbones for tiny action recognition.





\begin{figure}[t]
    \centering
    \includegraphics[scale=0.5]{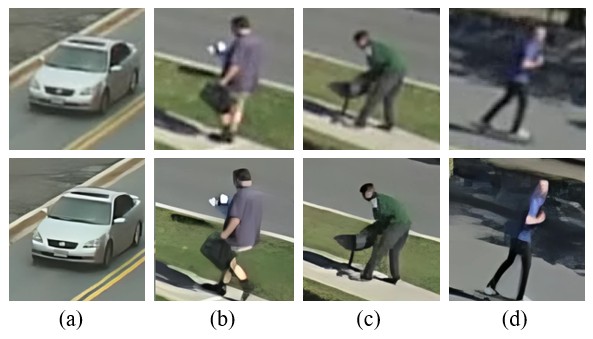}
    \caption{Super-Resolution Videos. 
    We use RealBasicVSR \cite{chan2021investigating} to transform the original low-resolution videos as the corresponding super-resolution ones.
    As expected,
    super-resolution videos tend to emphasize action details and reduce sensor noise. }
    \label{fig:super-res}
\end{figure}

\section{Dual-Resolution Distillation}
\label{optimize}

As mentioned before,
one key problem in this challenge is that,
low-resolution videos contain blurred appearance and motion of human activities.
To enhance action details,
we propose to transform these videos into super-resolution ones.
Note that,
there is no super-resolution ground truth for each low-resolution video.
Hence,
we use the popular RealBasicVSR \cite{chan2021investigating} as a fixed generator, 
and 
directly feed low-resolution videos to generate super-resolution ones with spatial size of 224$\times$224.
Since super-resolution training videos inherit activity labels from their original low-resolution ones,
we can use them to train video backbones for activity recognition.

In this work,
we propose a dual-resolution distillation framework,
which aims at leveraging super-resolution action knowledge to boost low-resolution recognition.
First,
we train a video backbone by using super-resolution videos. 
Second,
we use this recognition network as knowledge extractor,
i.e.,
for each low-resolution training video,
we feed its corresponding super-resolution video into extractor,
and generate the prediction score vector of activity categories as knowledge $\mathbf{k}$.
Finally,
we feed this low-resolution video into a low-resolution recognition network.
For training this network, 
we use two supervision signals on the prediction vector $\mathbf{p}$,
where
$\mathcal{L}_{bce}$ is the binary cross entropy loss between the prediction vector and ground truth label.
$\mathcal{L}_{kd}$ is the knowledge distillation loss (e.g., MSE) between the prediction vector $\mathbf{p}$ and the super-resolution knowledge vector $\mathbf{k}$,
\begin{equation}
\begin{aligned}
\mathcal{L}_{total} 
&=\alpha\mathcal{L}_{bce} +(1-\alpha)\mathcal{L}_{kd},\\
\mathcal{L}_{kd}
&=\frac{1}{C}\sum\nolimits_{c=1}^{C} (\mathbf{p}_{c}-\mathbf{k}_{c})^2,
\label{KD}
\end{aligned}
\end{equation}
where $\alpha$ is a weight scale for loss balance.
Our dual-resolution distillation flexibly leverages  super-resolution information to boost low-resolution recognition,
and thus leads to better performance as shown in Table \ref{tab:distillation}.

\begin{table}[t]
\small
\centering
\begin{tabular}{lc}
\toprule
ir-CSN                            & F1 Score       \\ \midrule
Without            & 0.403          \\
Uniform Sample   & 0.442          \\
Uniform Sample + Data Balance         & \textbf{0.469} \\ \midrule
UniFormer                            & F1 Score       \\ \midrule
Without     & 0.404          \\
Uniform Sample & 0.439          \\
Uniform Sample + Data Balance         & \textbf{0.452} \\ \bottomrule
\end{tabular}
\caption{Video Backbones with Key Training Settings.}
\label{tab:training}
\end{table}

\begin{table}[t]
\small
\centering
\begin{tabular}{lc}
\toprule
Model                                                     & F1 Score       \\ \midrule
2021 TinyAction Top 1 Model \cite{tirupattur2021tinyaction}     & 0.478          \\
ir-CSN                                                    & 0.469          \\
ir-CSN (SR)                                            & 0.484          \\
ir-CSN (SR+KD)                                         & \textbf{0.492} \\ \bottomrule
\end{tabular}
\caption{Dual-Resolution Distillation. SR: We use super-resolution videos to train a video backbone. SR+KD: We further perform dual-resolution distillation by the SR-trained backbone.}
\label{tab:distillation}
\end{table}

\begin{table*}[t]
\small
\centering
\begin{tabular}{lcc}
\toprule
Model Ensemble & Postprocessing & F1 Score  \\ \midrule
UniFormer$\times$1 
+ UniFormer(SR)$\times$2 
+ ir-CSN$\times$4 
+ ir-CSN(SR)$\times$3 
+ ir-CSN(SR+KD)$\times$2 
& Yes    & \textbf{0.883} \\ 
\bottomrule
\end{tabular}
\caption{Model Ensemble.}
\label{tab:ensamble}
\end{table*}

\section{Model Ensemble}
\label{sec:Post Processing}

We perform model ensemble to make final prediction.
We find that,
the models trained at the middle epochs are also useful to improve recognition.
Hence,
we choose 12 networks in total for model ensemble,
e.g.,
ir-CSN$\times$4 in Table \ref{tab:ensamble} means that, 
we choose 4 ir-CSN models which are trained at different epochs.
Furthermore,
due to long-tailed distribution in this challenge,
we carefully set the prediction thresholds for all the activity categories,
i.e.,
large-sample-size categories need a higher prediction threshold,
while
small-sample-size categories need a lower prediction threshold.
Finally,
we use the prior knowledge of categories \cite{tirupattur2021tinyaction} to assist activity recognition in the process of model ensemble,
i.e.,
within the same group of activities categories, 
it is reasonable to keep one class with the highest score. 
As shown in Table \ref{tab:ensamble},
we achieve a quite good F1 score by model ensemble and post-processing.

\section{Training Practices}
\label{experience}
We mainly use binary cross-entropy loss to deal with multi-label classification.
Additionally,
we also use Asymmetric Loss(ASL) \cite{ben2020asymmetric}, 
which is based on the Focal Loss,
in order to further relieve the unbalanced distribution problem of this multi-label benchmark in the challenge. 
AdamW is chosen as optimizer,
and warmup mechanism is used at the start phase of training. 
The initial learning rate for UniFormer is 2e-4,
and this rate for ir-CSN is 1e-4. 
We use Cosine Annealing Warm Restarts method to schedule the learning rate. 
The drop path rate of UniFormer is 0.4, 
and the dropout rate for ir-CSN is 0.5.

\section{Conclusion}
\label{conclusion}
In this work,
we design a comprehensive solution for low-resolution activity recognition in Tiny Actions Challenge.
First,
we choose effective video backbones with data balance training.
Second,
we design a dual-distillation framework to enhance action clues from super-resolution knowledge.
Finally,
we perform model ensemble and post-processing to further boost recognition on such long-tailed categories in this challenge.


\section{Acknowledgement}
This work is partially supported by 
the National Natural Science Foundation of China (61876176,U1813218), 
the Joint Lab of CAS-HK, 
the Shenzhen Research Program (RCJC20200714114557087),  
the Shanghai Committee of Science and Technology, China (Grant No. 21DZ1100100), 
Shenzhen Institute of Artificial Intelligence and Robotics for Society.
Thanks for Kunchang Li on valuable discussions about UniFormer and video classification in general.

{\small
\bibliographystyle{ieee_fullname}
\bibliography{egbib}

@String(ICPR = {Int. Conf. Pattern Recog.})

@String(ICPR  = {ICPR})

@article{chen2025videochat,
  title={VideoChat-M1: Collaborative Policy Planning for Video Understanding via Multi-Agent Reinforcement Learning},
  author={Chen, Boyu and Wang, Zikang and Yue, Zhengrong and Yan, Kainan and Yu, Chenyun and Huang, Yi and Liu, Zijun and Wen, Yafei and Chen, Xiaoxin and Liu, Yang and others},
  journal={arXiv preprint arXiv:2511.19524},
  year={2025}
}

@article{chen2025top,
  title={When Top-ranked Recommendations Fail: Modeling Multi-Granular Negative Feedback for Explainable and Robust Video Recommendation},
  author={Chen, Siran and Chen, Boyu and Yu, Chenyun and Ouyang, Yi and Lei, Cheng and Zhuo, Chengxiang and Li, Zang and Wang, Yali},
  journal={arXiv preprint arXiv:2511.18700},
  year={2025}
}

@article{chen2025g,
  title={G-UBS: Towards Robust Understanding of Implicit Feedback via Group-Aware User Behavior Simulation},
  author={Chen, Boyu and Chen, Siran and Yue, Zhengrong and Yan, Kainan and Yu, Chenyun and Kong, Beibei and Lei, Cheng and Zhuo, Chengxiang and Li, Zang and Wang, Yali},
  journal={arXiv preprint arXiv:2508.05709},
  year={2025}
}

@article{yue2025uniflow,
  title={UniFlow: A Unified Pixel Flow Tokenizer for Visual Understanding and Generation},
  author={Yue, Zhengrong and Zhang, Haiyu and Zeng, Xiangyu and Chen, Boyu and Wang, Chenting and Zhuang, Shaobin and Dong, Lu and Du, KunPeng and Wang, Yi and Wang, Limin and others},
  journal={arXiv preprint arXiv:2510.10575},
  year={2025}
}

@article{chen2025super,
  title={Super Encoding Network: Recursive Association of Multi-Modal Encoders for Video Understanding},
  author={Chen, Boyu and Chen, Siran and Li, Kunchang and Xu, Qinglin and Qiao, Yu and Wang, Yali},
  journal={arXiv preprint arXiv:2506.07576},
  year={2025}
}

@article{wang2025videochat,
  title={VideoChat-A1: Thinking with Long Videos by Chain-of-Shot Reasoning},
  author={Wang, Zikang and Chen, Boyu and Yue, Zhengrong and Wang, Yi and Qiao, Yu and Wang, Limin and Wang, Yali},
  journal={arXiv preprint arXiv:2506.06097},
  year={2025}
}

@article{chen2025vragent,
  title={VRAgent-R1: Boosting Video Recommendation with MLLM-based Agents via Reinforcement Learning},
  author={Chen, Siran and Chen, Boyu and Yu, Chenyun and Luo, Yuxiao and Yi, Ouyang and Cheng, Lei and Zhuo, Chengxiang and Li, Zang and Wang, Yali},
  journal={arXiv preprint arXiv:2507.02626},
  year={2025}
}

@article{chen2025percept,
  title={Percept, Chat, Adapt: Knowledge transfer of foundation models for open-world video recognition},
  author={Chen, Boyu and Chen, Siran and Li, Kunchang and Xu, Qinglin and Qiao, Yu and Wang, Yali},
  journal={Pattern Recognition},
  volume={160},
  pages={111189},
  year={2025},
  publisher={Elsevier}
}

@article{chen2025lvagent,
  title={Lvagent: Long video understanding by multi-round dynamical collaboration of mllm agents},
  author={Chen, Boyu and Yue, Zhengrong and Chen, Siran and Wang, Zikang and Liu, Yang and Li, Peng and Wang, Yali},
  journal={arXiv preprint arXiv:2503.10200},
  year={2025}
}

@inproceedings{tran2019video,
  title={Video classification with channel-separated convolutional networks},
  author={Tran, Du and Wang, Heng and Torresani, Lorenzo and Feiszli, Matt},
  booktitle={Proceedings of the IEEE/CVF International Conference on Computer Vision},
  pages={5552--5561},
  year={2019}
}

@article{chan2021investigating,
  title={Investigating Tradeoffs in Real-World Video Super-Resolution},
  author={Chan, Kelvin CK and Zhou, Shangchen and Xu, Xiangyu and Loy, Chen Change},
  journal={arXiv preprint arXiv:2111.12704},
  year={2021}
}

@article{tirupattur2021tinyaction,
  title={TinyAction Challenge: Recognizing Real-world Low-resolution Activities in Videos},
  author={Tirupattur, Praveen and Rana, Aayush J and Sangam, Tushar and Vyas, Shruti and Rawat, Yogesh S and Shah, Mubarak},
  journal={arXiv preprint arXiv:2107.11494},
  year={2021}
}

@inproceedings{demir2021tinyvirat,
  title={Tinyvirat: Low-resolution video action recognition},
  author={Demir, Ugur and Rawat, Yogesh S and Shah, Mubarak},
  booktitle={2020 25th International Conference on Pattern Recognition (ICPR)},
  pages={7387--7394},
  year={2021},
  organization={IEEE}
}

@article{ben2020asymmetric,
  title={Asymmetric loss for multi-label classification},
  author={Ben-Baruch, Emanuel and Ridnik, Tal and Zamir, Nadav and Noy, Asaf and Friedman, Itamar and Protter, Matan and Zelnik-Manor, Lihi},
  journal={arXiv preprint arXiv:2009.14119},
  year={2020}
}

@inproceedings{
li2022uniformer,
title={UniFormer: Unified Transformer for Efficient Spatial-Temporal Representation Learning},
author={Kunchang Li and Yali Wang and Gao Peng and Guanglu Song and Yu Liu and Hongsheng Li and Yu Qiao},
booktitle={International Conference on Learning Representations},
year={2022},
url={https://openreview.net/forum?id=nBU_u6DLvoK}
}
}

\end{document}